\documentclass[10pt,a4paper,twocolumn]{article}
\usepackage[T1]{fontenc}
\usepackage[utf8]{inputenc}
\usepackage{lmodern}
\usepackage{geometry}
\usepackage{amsmath,amssymb,amsthm,mathtools}
\usepackage{booktabs,array,tabularx,multirow}
\usepackage{graphicx,microtype,url}
\usepackage{tikz}
\usetikzlibrary{arrows.meta,positioning,fit,calc,shapes.geometric}
\usepackage[numbers,sort&compress]{natbib}
\usepackage[hidelinks]{hyperref}
\usepackage{enumitem}
\usepackage{caption}
\usepackage{setspace}
\newtheorem{definition}{Definition}

\newtheorem{proposition}{Proposition}

\newcommand{\E}{\mathbb{E}}
\newcommand{\Prob}{\mathbb{P}}
\newcommand{\1}{\mathbb{I}}

\newcommand{\CER}{\operatorname{CER}}
\newcommand{\WER}{\operatorname{WER}}

\newcommand{\Risk}{\operatorname{Risk}}

\newcommand{\IoU}{\operatorname{IoU}}
\newcommand{\keywords}[1]{\par\noindent\textbf{Keywords:} #1\par\vspace{0.5em}}
\title{Human-Centred, Risk-Constrained Evaluation of Financial Document Recognition: Evidence, Critical Errors, and Selective Automation}
\author{Subodh Lonkar\\
Independent Researcher}
\date{}
\begin{document}
\maketitle

\begin{abstract}
Financial document recognition is often assessed using character and word error rates, although these metrics do not express whether a system preserves the financial facts on which people and organisations act. This paper develops a human-centred, risk-constrained framework for evaluating financial document recognition as evidence recovery and decision support. It defines typed financial facts linked to visible source regions and separates verbatim transcription, deterministic normalization, and business interpretation. A layered quality model covers text, localization, layout, tables, fields, fact correctness, evidence grounding, uncertainty, robustness, and operational performance. Non-compensatory acceptance gates use one-sided uncertainty bounds for critical-field errors, preventing strong aggregate scores from masking failures in amounts, identifiers, signs, periods, or entity associations. Selective automation is formulated as a constrained routing problem that accounts for residual risk, human-review capacity, and review cost. The framework further specifies annotation and adjudication procedures, leakage-resistant benchmark partitions, statistical estimands, subgroup analyses, and a reproducibility card. Formal results establish the limitations of compensatory scoring, the non-zero upper risk bound after zero observed errors, and a loss-sensitive decision rule for routing uncertain cases to human review. The contribution is theoretical and methodological; no new model, dataset, or empirical performance result is claimed.
\end{abstract}


\section{Introduction}
Financial institutions and enterprises process documents whose content is both visually heterogeneous and semantically consequential. A bank statement may contain transaction dates, descriptions, debit and credit columns, balances, account identifiers, and running-balance relationships. An invoice may contain supplier identity, tax identifiers, line items, tax rates, discounts, currency, payment terms, and a payable total. A financial report may combine narrative disclosures, multi-level tables, footnotes, units, comparative periods, and negative values represented with parentheses. OCR is frequently the first machine-readable transformation in these workflows.

The common practice of reporting a single character error rate (CER) or word error rate (WER) is insufficient for this domain. Edit-distance metrics treat all character substitutions similarly, whereas the operational impact of an error depends on its semantic role and downstream use. Confusing a letter in a narrative paragraph is not equivalent to changing `1,250.00' to `12,500.00', dropping a minus sign, shifting a value from the debit column to the credit column, or associating a tax amount with the wrong invoice. Document understanding research has established the importance of layout, relations, and visual context in addition to text recognition \cite{jaume2019,mathew2021,huang2022,kim2022,li2023,udop2023,docllm2024}. Recent financial-specific work has also highlighted fact-level evaluation for numerical, temporal, entity, and financial-concept fidelity \cite{he2025}.

This paper argues that financial OCR should be evaluated as an evidence-recovery system, not merely a text transcription engine. A system is useful only to the extent that it faithfully recovers information visible in the source, preserves the structure that gives the information meaning, expresses uncertainty when evidence is inadequate, and supports safe downstream processing. The proposed framework therefore evaluates a sequence of increasingly semantic capabilities while retaining component-level diagnostics.

\subsection{Research questions}
The framework is designed to answer six questions:
\begin{enumerate}[leftmargin=*]
\item How accurately does a system detect and transcribe all visible text, including financial symbols and low-frequency identifiers?
\item Does it preserve spatial and logical structure, particularly tables, merged headers, row associations, and reading order?
\item Does it recover typed financial fields with correct values, units, signs, dates, currencies, and entity associations?
\item Can it distinguish what is visually supported from what is inferred, normalised, or unavailable?
\item Are its confidence estimates meaningful enough to support abstention, escalation, and human review?
\item Does its quality remain acceptable across document categories, acquisition conditions, languages, institutions, and operational loads?
\end{enumerate}

\subsection{Contributions and scope}
The paper makes seven methodological contributions. First, it specifies a typed, evidence-linked representation of financial facts that separates visible evidence from deterministic normalization and business interpretation. Second, it defines a layered quality vector and risk functional that retain diagnostic information rather than collapsing errors into a lexical score. Third, it formalizes non-compensatory acceptance gates using prespecified critical-field constraints and uncertainty bounds. Fourth, it models selective automation as a human-centred routing decision that accounts for automated risk, human-review effectiveness, review cost, and review capacity. Fifth, it specifies the evidence and explanation packet required for a reviewer to verify, correct, or escalate a prediction. Sixth, it provides an annotation, sampling, statistical-reporting, and reproducibility protocol. Seventh, it translates the framework into a production control-plane blueprint with versioned interfaces, fail-safe states, staged rollout, audit events, monitoring, and requalification. The four synthetic document exemplars make the evaluation units and cross-document checks concrete but are not an empirical dataset. These are theoretical formalizations and design principles, not claims of a new OCR model or an empirically validated benchmark. The framework applies to OCR-dependent pipelines and OCR-free document models; both must provide output that can be evaluated against the same visual evidence contract \cite{kim2022}.

The paper does not prescribe one universal weighting scheme. Weights, criticality classes, tolerance thresholds, and acceptance limits are application risk decisions that must be fixed before test results are examined. The framework also does not claim that a benchmark score certifies regulatory compliance, fraud detection, or correctness on all future documents.

\subsection{Human-centred decision context}
The intended decision-maker is not an abstract automated pipeline but a person or team accountable for a downstream workflow. The framework therefore treats evidence presentation, uncertainty communication, review capacity, override authority, and auditability as evaluation concerns. A system output should be accompanied by a compact evidence packet containing the predicted value, the verbatim visual span, the source-region locator, the normalization rule (if any), the relevant table row/column or entity association, confidence at the decision unit, and any consistency-check result. Reviewers should be able to inspect the source image and reject the system's interpretation without losing the original evidence.

Human review is not assumed to be infallible or costless. The reviewer may correct, confirm, or escalate a prediction; the workflow should record the action, time, reason code, and any subsequent correction. Evaluation should distinguish (i) machine error detected and corrected by a reviewer, (ii) machine error accepted by a reviewer, (iii) correct machine output unnecessarily changed, and (iv) unresolved ambiguity escalated for adjudication. This makes human oversight measurable rather than treating the presence of a human checkpoint as sufficient evidence of safety. The framework does not prescribe a particular user-interface design; it specifies the information and outcomes that an interface must make assessable.

\section{Financial Document Landscape and Evaluation Units}
\subsection{Document taxonomy}
Financial data is not one homogeneous OCR domain. The benchmark should explicitly represent document families and their distinguishing visual and semantic characteristics. Table~\ref{tab:categories} proposes a taxonomy. Categories may overlap: for example, an invoice can be scanned, multilingual, handwritten, and table-heavy.

\begin{table*}[!t]
\caption{Financial document categories and category-specific evaluation emphasis}
\label{tab:categories}
\centering\footnotesize
\begin{tabularx}{\textwidth}{p{0.16\textwidth}p{0.27\textwidth}X}
\toprule
Category & Representative content & High-priority evaluation dimensions \\
\midrule
Bank statements & Account identifiers, opening/closing balances, transaction dates, descriptions, debit/credit, running balances & Sign and decimal fidelity; row-to-column association; transaction completeness; balance continuity; date and amount normalization \\
Invoices and credit notes & Supplier/buyer, invoice number, line items, tax, discounts, totals, payment terms & Key-field exact match; line-item table structure; tax arithmetic; currency and unit fidelity; credit-note sign semantics \\
Receipts and expense claims & Merchant, date/time, items, subtotal, tax, tip, total, payment method & Low-resolution robustness; total extraction; item/quantity association; locale-specific separators; cropped or curved images \\
Purchase orders and remittance advices & PO references, quantities, unit prices, remitted amounts, invoice references & Cross-document identifier matching; row alignment; partial payment semantics; quantity and unit precision \\
Tax and payroll documents & Taxpayer/employer IDs, periods, gross/net pay, deductions, rates, statutory fields & Identifier exactness; period and jurisdiction; negative/withheld values; sensitive-field completeness and access controls \\
Loan, credit, and insurance documents & Principal, interest rate, repayment schedule, fees, collateral, coverage, dates & Decimal and percentage fidelity; schedule/table reconstruction; term association; unit and effective-date semantics \\
Trade finance & Bills of lading, letters of credit, packing lists, certificates, invoices & Multi-document entity consistency; quantities and units; dates; reference numbers; stamps and handwriting \\
Financial statements and regulatory disclosures & Balance sheets, income/cash-flow statements, notes, comparative tables, footnotes & Multi-level headers; signs and scale (thousands/millions); period alignment; footnote linkage; cross-page table continuation \\
Market and investment documents & Contract notes, statements, holdings, corporate actions, confirmations & Instrument identifiers; quantity/price; currency; event dates; percentage and basis-point precision \\
\bottomrule
\end{tabularx}
\end{table*}

\subsection{Evaluation units and hierarchy}
A benchmark must retain the hierarchy of the source material. Let a corpus contain documents $d\in\mathcal D$, pages $p\in\mathcal P_d$, regions $r\in\mathcal R_{dp}$, and semantic fields or facts $f\in\mathcal F_d$. Each field is associated with a source region or set of regions, a field type, a canonical value where applicable, and a criticality class. The hierarchy matters statistically: pages from one document, statements from one account, or invoices from one supplier are not necessarily independent observations.

\begin{definition}[Evidence-grounded annotation]
An annotation for field $f$ is evidence-grounded if it records (a) the exact visible source span or region, (b) the verbatim transcription when text is visible, (c) a typed canonical value when normalization is permitted, (d) the semantic role and unit, and (e) an explicit status for illegible, absent, ambiguous, or not-applicable content.
\end{definition}

Verbatim transcription and canonical interpretation must be stored separately. For example, the visible string `1.250,00 EUR' may be transcribed exactly and separately normalized to a decimal amount and ISO currency code under a declared locale convention. Normalization must not silently repair uncertain source text. The annotation should preserve provenance so that evaluators can distinguish recognition from interpretation.

\subsection{Existing benchmark resources and coverage gaps}
No single public dataset spans the full financial OCR problem. Table~\ref{tab:datasets} summarizes complementary resources and the role they can play in a benchmark portfolio. These datasets should not be treated as interchangeable or assumed to provide all required annotations. General document-parsing and multimodal benchmarks, including OmniDocBench, OCRBench, OCRBench v2, and MMLongBench-Doc, extend coverage of layout, visual text, and long-context evidence, but do not automatically provide financial-workflow risk labels or human-review outcomes \cite{ouyang2025omni,liu2023ocrbench,fu2025ocrbenchv2,ma2024mmlong}. In particular, a text-only financial reasoning dataset is not an OCR test set, and a table-only dataset does not evaluate document-level field association or operational reliability.

\begin{table*}[!t]
\caption{Representative public datasets and their appropriate role in a financial OCR evaluation portfolio}
\label{tab:datasets}
\centering\footnotesize
\begin{tabularx}{\textwidth}{p{0.17\textwidth}p{0.20\textwidth}p{0.29\textwidth}X}
\toprule
Resource & Domain / modality & Useful annotations or tasks & Important limitation for the proposed evaluation \\
\midrule
SROIE \cite{huang2019} & Scanned receipts & Text localization, OCR, key information extraction & Receipt-focused; does not represent the full range of statements, disclosures, or multi-page finance documents \\
CORD \cite{park2019} & Indonesian receipts & Word boxes, text, hierarchical semantic receipt labels & Strong receipt parsing resource, but narrow document and language coverage \\
FUNSD \cite{jaume2019} & Noisy scanned forms & Text, entities, semantic relations, layout & General forms rather than a finance-specific corpus \\
DocVQA \cite{mathew2021} & General document images & Document visual question answering with image-grounded questions & QA accuracy is not a direct substitute for exhaustive OCR transcription \\
FinTabNet / GTE \cite{zheng2021} & Financial-report tables & Table detection and cell-structure annotations & Table-centric; does not alone test full-page reading order, field extraction, or operational routing \\
PubTables-1M \cite{smock2022} & Scientific-document tables & Table structure, cells, headers, canonicalized structure & Valuable for structural methods, but its domain differs from financial statements \\
FinQA \cite{chen2021finqa} & Financial reports, text and tables & Numerical reasoning questions with supporting facts and programs & Primarily a reasoning dataset; source text availability does not make it an image OCR benchmark \\
FinCriticalED \cite{he2025} & Financial document images & Expert-annotated financial facts and context-sensitive visual evaluation & Directly relevant; still one benchmark resource and not a complete substitute for workflow, calibration, robustness, and cost evaluation \\
SynFinTabs \cite{bradley2024} & Synthetic financial tables & Table images with structured ground truth and spatial annotations & Useful for controlled stress testing; synthetic distributions may not match naturally occurring documents \\
\bottomrule
\end{tabularx}
\end{table*}

A robust benchmark should combine these public resources where their licenses and annotation schemas permit, and supplement them with a governed, representative corpus of real financial documents. The real-world corpus should include multiple institutions, vendors, jurisdictions, languages, and acquisition channels. Public data should be used for reproducibility and component diagnostics; enterprise data is needed to estimate performance on the actual target population. Dataset licenses, privacy conditions, redistribution rights, and source-document permissions must be checked independently before release or commercial use.

\section{Recent Document AI and Evaluation Literature}
Recent document AI systems increasingly combine visual, textual, and spatial representations, while some end-to-end models generate structured outputs without a conventional OCR stage. Examples include unified vision--text--layout pretraining, OCR-free document understanding, and layout-aware generative models \cite{huang2022,kim2022,li2023,udop2023,docllm2024}. These developments strengthen the case for evaluating systems against a common evidence contract rather than assuming that a particular internal pipeline is the unit of comparison. Recent surveys document the breadth of scanned-form understanding methods and the expansion of multimodal visually rich document understanding \cite{abdallah2024,ding2026}.

Benchmark design has also moved beyond isolated character recognition. OCRBench and OCRBench v2 cover broad text-centric visual tasks, while MMLongBench-Doc tests long-context document understanding and cross-page evidence use \cite{liu2023ocrbench,fu2025ocrbenchv2,ma2024mmlong}. OmniDocBench provides multi-level document parsing annotations across diverse document sources, and TableFormer illustrates the importance of explicit table-structure modelling \cite{ouyang2025omni,nassar2022}. These resources are useful for component evaluation, but their coverage does not by itself establish financial fact risk, human-review effectiveness, or deployment-specific acceptance limits.

The financial-document setting has a particularly relevant emerging benchmark in FinCriticalED, which evaluates visually grounded financial facts and demonstrates why lexical similarity alone is inadequate \cite{he2025}. The present framework is complementary: it does not claim to replace that benchmark or reproduce its empirical findings. Instead, it formalizes a broader evaluation and governance layer around critical-field estimands, confidence bounds, clustered sampling, selective routing, reviewer outcomes, and operational constraints. For risk control, the framework draws on established calibration and selective-prediction methods and more recent conformal risk-control work \cite{guo2017,geifman2017,angelopoulos2024}.

\section{Reference Architecture and Output Contract}
The evaluation architecture separates the recognition system from an independent adapter and evaluator. It supports conventional OCR pipelines, layout-aware models, and OCR-free document models. Each system receives the same image and must produce a standardized output contract. An adapter converts native outputs into the contract without adding information that the system did not produce; the production control flow is summarized later in Fig.~\ref{fig:workflow}.

\begin{figure*}[!t]
\centering
\includegraphics[width=0.98\textwidth]{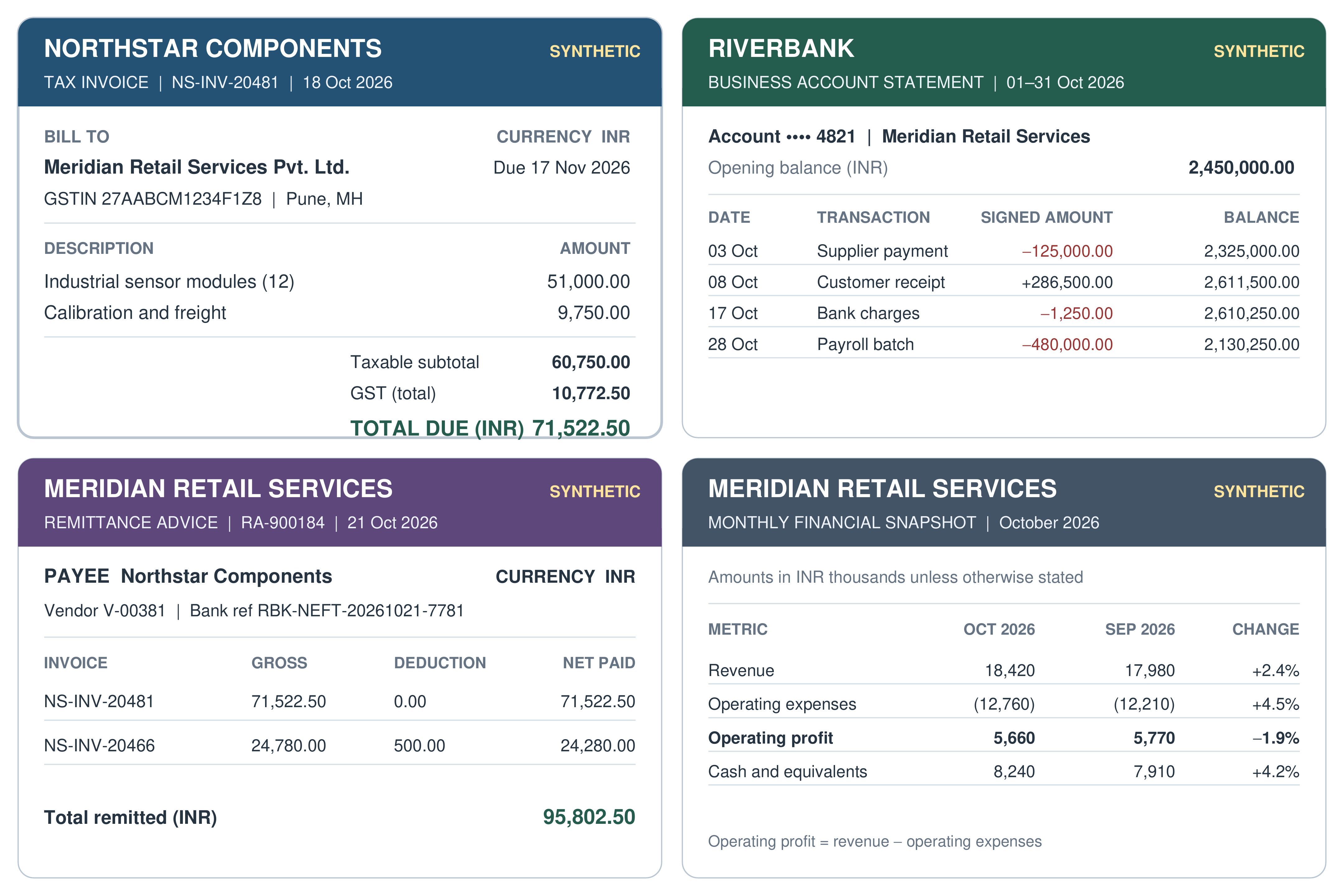}
\caption{Programmatically typeset synthetic financial-document exemplars: (a) supplier invoice, (b) bank statement, (c) remittance advice, and (d) monthly financial snapshot. All entities, identifiers, dates, and amounts are fictitious and were created solely to illustrate the proposed evaluation units using a deterministic Python/ReportLab script; these images are not benchmark observations or empirical evidence}
\label{fig:syntheticdocs}
\end{figure*}

A minimum output contract should include document and page identifiers; detected text spans; verbatim text; polygon or bounding box; reading-order index; table and cell identifiers; extracted field name and typed value; currency, unit, sign, and date interpretation where relevant; confidence or uncertainty if available; source-region references; and explicit abstention or error states. Missing confidence must be reported as unavailable, not imputed. An OCR-free model may emit fields directly, but it must still provide visual evidence references for auditable evaluation.

\subsection{Synthetic document exemplars and cross-document checks}
Figure~\ref{fig:syntheticdocs} supplies four programmatically typeset, fictitious document examples. Their purpose is to make the annotation contract concrete, not to estimate model performance. The invoice exposes supplier identity, purchase-order linkage, line-item amounts, tax, and payable total. The bank statement exposes transaction date, reference, debit/credit polarity, and running balance. The remittance advice exposes invoice-to-payment matching, deductions, and net settlement. The management snapshot exposes period association, units, negative values, and arithmetic relationships.

A production evaluation should represent each fact as a typed record with at least: \texttt{document\_id}, \texttt{page\_id}, \texttt{field\_path}, \texttt{verbatim\_text}, \texttt{canonical\_value}, \texttt{data\_type}, \texttt{currency\_or\_unit}, \texttt{period}, \texttt{polarity}, \texttt{entity\_key}, \texttt{evidence\_polygon}, \texttt{normalization\_rule\_id}, \texttt{model\_version}, and \texttt{confidence}. The record should also carry a validation status and an explicit state for missing, illegible, ambiguous, or not-applicable values. A missing confidence score is not a zero confidence score; it is unavailable and must be handled by policy.

The examples illustrate why evaluation should include both within-document and cross-document relations. In the statement, the payroll transaction is shown in parentheses in the debit column. If a system emits a positive signed amount, the error is not merely a character mismatch: it is a polarity error that can also break the running-balance relation. In the invoice/remittance pair, the invoice total should be reconciled to the remittance gross amount before deductions; the net paid amount is a different semantic quantity. In the monthly snapshot, parentheses denote negative expenses and the operating-profit identity is a consistency check, not a substitute for reading the displayed values. These checks should be reported as separate validator outcomes so that a coincidental arithmetic match cannot conceal an unsupported extraction.

For an illustrative deployment, an evidence packet for a payroll transaction would contain the source crop, polygon coordinates, the verbatim string ``(480,000.00)'', the normalized signed value, the debit-column association, the running-balance context, the model confidence, and the validator result. A reviewer can then distinguish an OCR transcription error from a polarity-normalization error or a table-association error. This explicit provenance is the operational bridge between a benchmark score and a reviewable production decision.

\subsection{Separation of recognition and normalization}
The evaluator should score at least three representations:
\begin{enumerate}[leftmargin=*]
\item \textit{Visual transcript}: characters and tokens as visibly printed, preserving punctuation, signs, separators, and symbols.
\item \textit{Canonical field}: typed representation after deterministic, declared normalization, such as date parsing or currency normalization.
\item \textit{Business interpretation}: a semantic relation, such as whether an amount is debit, credit, tax, gross pay, or closing balance.
\end{enumerate}
A system should not receive credit for a correct canonical value if the visible transcription is wrong in a way that matters for auditability, nor should benign formatting differences be counted as semantic failures when the canonical value is equivalent. Both levels should be reported.

\section{A Multi-Layer Quality Model}
I define quality as a vector rather than a single primitive score:
\begin{equation}
\begin{aligned}\mathbf Q(s)=(&Q_{\rm det},Q_{\rm txt},Q_{\rm lay},Q_{\rm tab},Q_{\rm fld},\\&Q_{\rm fact},Q_{\rm evd},Q_{\rm cal},Q_{\rm rob},Q_{\rm ops}).\end{aligned},
\label{eq:qvector}
\end{equation}
where $s$ denotes a system. The components measure text-region detection, transcription, layout and reading order, table structure, typed field extraction, financial fact correctness, evidence grounding, uncertainty calibration, robustness, and operational performance. Reporting this vector prevents a strong result on one layer from concealing a weakness in another.

\subsection{Text recognition: CER and WER}
For a reference string $y$ and hypothesis $\hat y$, let $S$, $D$, and $I$ denote the minimum number of substitutions, deletions, and insertions under Levenshtein alignment \cite{levenshtein1966}. Then
\begin{equation}
\CER=\frac{S_c+D_c+I_c}{N_c},\qquad
\WER=\frac{S_w+D_w+I_w}{N_w},
\label{eq:cerwer}
\end{equation}
where $N_c$ and $N_w$ are the numbers of reference characters and words, respectively. Corpus-level micro-averaged CER is total edit operations divided by total reference characters; macro-averaged document CER is the arithmetic mean of per-document CERs. Both should be reported because micro-averaging gives long documents greater influence. The tokenization policy, Unicode normalization, whitespace handling, and treatment of punctuation must be fixed in advance. CER and WER remain diagnostic metrics, not sufficient measures of financial correctness \cite{neudecker2021}.

\subsection{Detection and localization}
For ground-truth regions $G$ and predicted regions $P$, define intersection-over-union as
\begin{equation}
\IoU(P,G)=\frac{|P\cap G|}{|P\cup G|}.
\end{equation}
Region detection can be evaluated using precision, recall, and F1 at prespecified IoU thresholds. A matched region should also be scored for transcription. Detection-only scores can otherwise reward boxes that cover text without correctly reading it. Report localization metrics separately for text lines, words, tables, and non-text financial marks such as stamps or handwritten annotations.

\subsection{Layout and reading order}
Let $E^*$ be the set of precedence relations in the annotated reading-order graph and $\hat E$ the predicted relations. A pairwise reading-order F1 can be calculated from precision and recall over ordered region pairs. For multi-column statements and reports, this is more informative than comparing a single flattened transcript. Layout classes should include title, paragraph, table, header, footer, footnote, signature, stamp, and marginal note where applicable. Layout-aware document benchmarks motivate evaluating both spatial and semantic structure \cite{jaume2019,huang2022}.

\subsection{Table reconstruction}
A table is represented as a grid graph with cells, row and column indices, spans, headers, and cell text. Let $C^*$ and $\hat C$ be the ground-truth and predicted cell sets after canonicalization and matching. Report cell detection precision/recall/F1, row/column boundary accuracy, span accuracy, header association accuracy, and cell-text accuracy. A tree-edit or structure-similarity measure may supplement these metrics, but must not replace exact cell-level evaluation for critical amounts. Table extraction datasets and models illustrate the importance of canonicalized structural annotations and cell-level consistency \cite{smock2022,zheng2021,nassar2022,ouyang2025omni}; these resources remain complementary to financial fact-level evaluation.

\subsection{Typed field extraction}
For each field type $k$, define exact canonical match:
\begin{equation}
M_{f}=\1\{\operatorname{canon}(\hat v_f)=\operatorname{canon}(v_f^*)\}.
\end{equation}
The canonicalization function must be type-specific and versioned. For amounts, equality should be evaluated using decimal arithmetic at a declared scale, not binary floating-point tolerance chosen after observing errors. For identifiers, exact string match is generally appropriate after only explicitly permitted normalization. For dates, equivalence may be evaluated across declared formats only if day/month ambiguity is resolved by metadata or source convention. Report field precision, recall, F1, exact-match accuracy, and missing/extra-field rates. Macro-average across field types and micro-average across instances.

\subsection{Financial fact correctness}
A financial fact is a typed tuple, for example
\begin{equation}
\begin{aligned}
f=(&\text{entity},\text{attribute},\text{value},\text{unit},\\
   &\text{period},\text{polarity},\text{evidence}).
\end{aligned}
\end{equation}
Fact correctness requires agreement on all applicable components. A value match without the correct unit, period, entity, or polarity is not a correct fact. This representation captures common high-impact failures: amount-scale mismatch, date-period shift, debit/credit inversion, incorrect reporting entity, and detached footnote. Financial fact-level evaluation is consistent with the direction of recent financial OCR benchmarking \cite{he2025}.

\section{Risk-Aware Error Measurement}
\subsection{Criticality classes}
Each field type is assigned a criticality class before evaluation. One possible taxonomy is: C0, cosmetic/non-semantic; C1, low-impact descriptive; C2, operationally relevant; C3, financially material or identity-critical; and C4, safety-, compliance-, or legally critical. The class depends on the intended workflow, not solely on the field name. For example, a date may be C2 in a search index but C4 when it determines a filing deadline. A customer identifier may be C4 in payment routing.

Let $w_f>0$ be a prespecified severity weight for field $f$, and let $e_f\in[0,1]$ be a normalized error indicator or loss. A weighted loss is
\begin{equation}
L_{\rm crit}=\frac{\sum_{f\in\mathcal F}w_f e_f}{\sum_{f\in\mathcal F}w_f}.
\label{eq:critical}
\end{equation}
This quantity is useful for ranking and monitoring, but it is compensatory: many correct low-risk fields can dilute a single critical error. Therefore, it must be accompanied by separate critical-error rates and hard acceptance gates.

\subsection{Critical error rate and severity-weighted risk}
Let $H_f=1$ indicate an error in a field designated critical under the workflow policy. The financial-critical error rate (FCR) is
\begin{equation}
FCR_{\rm crit}=\frac{\sum_f H_f\1\{\hat f\ne f^*\}}{\sum_f H_f},
\end{equation}
where a missing, extra, wrongly associated, or unsupported critical field is counted as an error; report the denominator, eligible-field coverage, and ambiguity exclusions explicitly. For document $d$, let $\ell_d$ be the application-defined loss associated with its OCR output. The expected operational loss is
\begin{equation}
\Risk(s)=\E_{d\sim\mathcal D}[\ell_d(s)].
\end{equation}
The loss function should be defined with business, control, and risk owners. It may incorporate severity, probability of downstream use, reversibility, and detectability. Monetary loss estimates should only be used where empirically supportable; otherwise report ordinal severity classes and event rates.

\subsection{Non-compensatory acceptance gates}
A system should not pass solely because its weighted average is high. Define gates for critical identifiers, monetary amounts, sign/polarity, currency/unit, reporting period, table association, and evidence grounding. A benchmark-level acceptance rule can be written:
\begin{equation}
\operatorname{Pass}(s)=\1\{Q_{\rm aggregate}\ge\tau\}
\prod_{g\in\mathcal G}\1\{U_g(s)\le\eta_g\},
\label{eq:gate}
\end{equation}
where $U_g$ is the measured error rate for gate $g$, $\eta_g$ is its prespecified maximum, and $\tau$ is a minimum aggregate threshold. In practice, the gate should use an upper confidence bound for error rates rather than only the point estimate. If a gate has zero observed errors, the upper confidence bound is not zero; sample size must be sufficient to support the desired assurance.

\subsection{Financial consistency checks}
Consistency checks are document-type-specific validators, not substitutes for visual evidence. Examples include:
\begin{align}
T&\approx S+\mathrm{Tax}+\mathrm{Fees}-\mathrm{Discount},\\
B_t&=B_{t-1}+C_t-D_t+J_t,\\
N&=G-\sum_i W_i+A,\\
B_{t+1}&=B_t+I_t-P_t.
\end{align}
Here $T$ is invoice total, $S$ subtotal, $B_t$ running balance, $C_t$ credits, $D_t$ debits, $J_t$ adjustments, $N$ net pay, $G$ gross pay, $W_i$ employee-side withholdings/deductions, and $A$ employee-side additions or adjustments, $I_t$ accrued interest, and $P_t$ payment applied to principal and interest according to the schedule convention. These equations are illustrative; real documents may include rounding, tax-specific rules, fees, reversals, value dates, or nonstandard accounting conventions. A validator must encode the applicable convention and tolerance. A failed arithmetic check indicates inconsistency, not necessarily OCR error: the source document itself may be inconsistent. The evaluator should distinguish source inconsistency from system-induced inconsistency.

\section{Evidence Grounding, Hallucination, and Uncertainty}
\subsection{Evidence grounding}
For every extracted field, the system should identify the image region supporting the value. Define evidence coverage as the proportion of required extracted fields with a valid source-region link. Define evidence precision as the proportion of supplied links that overlap the correct annotated evidence region above a declared threshold. A field with no visible support should be marked absent, illegible, or inferred according to the output contract. A plausible value inferred from context is not a successful OCR transcription.

\subsection{Confidence calibration}
If a system provides confidence $p_i$ for correctness indicator $y_i\in\{0,1\}$, the Brier score is \cite{brier1950}
\begin{equation}
\operatorname{BS}=\frac{1}{N}\sum_{i=1}^N(p_i-y_i)^2.
\end{equation}
Expected calibration error (ECE), widely used in neural-network calibration analysis \cite{guo2017}, can be estimated by partitioning predictions into bins $B_m$:
\begin{equation}
\operatorname{ECE}=\sum_{m=1}^{M}\frac{|B_m|}{N}
\left|\operatorname{acc}(B_m)-\operatorname{conf}(B_m)\right|.
\end{equation}
ECE depends on binning and can conceal subgroup miscalibration; report reliability diagrams, Brier score, and calibration by criticality class and document category. Confidence must be evaluated at the unit where it is used operationally (token, field, table, or document). A high token confidence does not imply a high probability that a debit/credit relation is correct.

\subsection{Selective prediction and review routing}
An operational OCR system may abstain or route uncertain cases to human review. Following the selective-prediction perspective \cite{geifman2017}, let $c(x)\in[0,1]$ be a confidence score and accept predictions when $c(x)\ge t$. Define coverage and selective risk:
\begin{align}
\operatorname{Cov}(t)&=\Prob(c(X)\ge t),\\
\operatorname{Risk}(t)&=\E[\ell(X,\hat Y)\mid c(X)\ge t].
\end{align}
Plot risk against coverage and report the area under the risk--coverage curve with a clearly specified numerical integration rule. Also report the fraction routed to review, residual critical-error rate among auto-accepted items, review workload, and end-to-end latency. The operating threshold should be selected on a calibration set and evaluated once on a held-out test set. Selective accuracy is not meaningful unless coverage is reported alongside it.

\section{Formal Problem Definition and Theoretical Properties}
\subsection{Document population, outputs, and target estimands}
Let $X$ denote a document sampled from a target population $\mathcal{P}$, and let $s\in\mathcal{S}$ be a recognition system. The system produces an output $Z_s(X)$, which may contain text regions, table structures, typed fields, confidence values, and evidence links. Let $Y(X)$ denote the adjudicated reference annotation. I distinguish the observable image $X$, the reference facts $F^*(X)$, and the system's predicted facts $\widehat F_s(X)$. The distinction is important: the image is the evidence; the annotation is a measurement of that evidence; and the predicted fact is the system's claim.

For each field instance $i$, define a reference tuple
\begin{equation}
f_i^*=(e_i,a_i,v_i,u_i,t_i,p_i,r_i),
\end{equation}
where $e_i$ is the entity, $a_i$ the attribute, $v_i$ the verbatim visible value, $u_i$ the unit or currency, $t_i$ the applicable time or reporting period, $p_i$ the polarity or semantic role, and $r_i$ the supporting image region. A predicted tuple is correct only when all components required by the field schema agree under a declared equivalence relation $\equiv_k$. The relation may permit harmless formatting equivalence (for example, a declared date-format conversion) but must not silently equate semantically distinct values (for example, debit and credit, parentheses and positive values, or dollars and thousands of dollars).

Define the field loss $\ell_i(s;X)\in[0,1]$ using a field-type-specific loss function fixed before evaluation. For a critical field, a binary loss is usually appropriate: $\ell_i=1$ if the required fact is absent, incorrect, attached to the wrong entity/row, or unsupported by its cited evidence; otherwise $0$. For non-critical descriptive fields, graded losses may be used. The primary population estimand is
\begin{equation}
R_{\mathcal{P}}(s)=\E_{X\sim\mathcal{P}}\left[\frac{\sum_{i\in I(X)}w_i\ell_i(s;X)}{\sum_{i\in I(X)}w_i}\right],
\label{eq:populationrisk}
\end{equation}
where $I(X)$ is the set of in-scope field instances and $w_i$ are prespecified policy weights. This is a document-first estimand: each document contributes one normalized loss, preventing documents with many fields from automatically dominating. A field-first estimand can also be reported, but it answers a different question and must be labelled accordingly.

For a criticality class $c$, define the conditional critical error probability
\begin{equation}
p_c(s)=\Prob_{X\sim\mathcal{P},\ i\sim I_c(X)}\{\ell_i(s;X)=1\},
\label{eq:criticalprob}
\end{equation}
where the sampling of a field instance is specified explicitly (uniform over documents then fields, or uniform over all eligible fields). The sampling convention is part of the estimand, not an implementation detail.

\subsection{Why a single average cannot certify financial reliability}
\begin{proposition}[Compensatory masking]
For any weighted arithmetic score $S=\sum_{k=1}^K w_kq_k$ with $K\ge2$, $w_k>0$, $\sum_k w_k=1$, and $q_k\in[0,1]$, there exist two systems $A$ and $B$ such that $S(A)>S(B)$ while a designated critical dimension satisfies $q_j(A)<q_j(B)$.
\end{proposition}
\begin{proof}
Choose any $j$ with $w_j<1$. Set all non-critical dimensions of $A$ to $1$ and its critical dimension to $0$. Set all dimensions of $B$ to a common value $b$ with $0<b<1$. Then $S(A)=1-w_j$ and $S(B)=b$. Choosing $b<1-w_j$ gives $S(A)>S(B)$ although $q_j(A)=0<b=q_j(B)$. Thus the aggregate ordering does not preserve the ordering on a critical dimension.\end{proof}

This proposition does not imply that composite scores are useless. It establishes that a composite score cannot, by itself, certify a safety-relevant constraint. A score may rank systems after feasibility has been established; it should not determine feasibility.

\subsection{Risk-constrained acceptance}
Let $\mathcal{G}$ be a set of critical gates, such as amount exactness, identifier exactness, polarity, currency/unit, period association, table-cell association, and evidence validity. For each gate $g$, let $p_g(s)$ be its target-population error probability and let $\eta_g$ be the maximum tolerable rate chosen by the workflow owner. Let $U_g(s;\alpha_g)$ be a one-sided upper confidence bound with nominal non-coverage probability $\alpha_g$. The system is eligible for automatic processing only if
\begin{equation}
\operatorname{Eligible}(s)=\1\{S(s)\ge\tau\}\prod_{g\in\mathcal{G}}\1\{U_g(s;\alpha_g)\le\eta_g\}.
\label{eq:eligible}
\end{equation}
If simultaneous coverage across gates is required, the confidence procedure must control the family-wise non-coverage probability, for example through a prespecified Bonferroni allocation $\sum_g\alpha_g\le\alpha$ or a justified joint procedure. Reporting separate nominal 95\% bounds for many gates does not yield a joint 95\% guarantee.

\begin{proposition}[Zero observed errors do not imply a zero risk bound]
Under independent Bernoulli sampling of $n$ eligible critical field instances, with $0<\alpha<1$, if zero errors are observed, the exact one-sided $(1-\alpha)$ Clopper--Pearson upper confidence limit is
\begin{equation}
U=1-\alpha^{1/n}.
\end{equation}
Consequently, demonstrating $p\le\eta$ with zero observed errors requires
\begin{equation}
n\ge\frac{\log\alpha}{\log(1-\eta)}.
\end{equation}
\end{proposition}
\begin{proof}
For a binomial random variable with success interpreted as an error, the probability of observing zero errors at error probability $p$ is $(1-p)^n$. The one-sided exact upper limit solves $(1-U)^n=\alpha$, yielding the first expression. Rearranging $1-\alpha^{1/n}\le\eta$ gives the second.\end{proof}

For illustration only, if a policy requires a one-sided 95\% upper bound below $10^{-3}$ and the independent-trial assumptions hold, the zero-error sample requirement is approximately 2,995 critical instances. This is not a recommended universal sample size. Clustering, unequal inclusion probabilities, annotation uncertainty, and multiple gates increase the design requirements. In particular, treating pages from one statement or invoices from one supplier template as independent trials can materially understate uncertainty.

\subsection{Selective automation as a constrained decision problem}
Let $c_s(X)$ be a confidence score calibrated for the event that the complete critical fact set for document $X$ is correct. For threshold $t$, the auto-accept set is $A_t=\{X:c_s(X)\ge t\}$ and the review set is $R_t=\{X:c_s(X)<t\}$. Define coverage $\gamma(t)=\Prob(X\in A_t)$. Let $\ell_A(X)$ denote loss if the system output is auto-accepted, and let $\ell_H(X)$ denote residual loss after human review, including reviewer error and unresolved ambiguity. Let $C_H(X)$ be review cost. A workflow-level objective is
\begin{align}
J(t)=\E[&\1\{X\in A_t\}\ell_A(X)\\
       &+\1\{X\in R_t\}\{\ell_H(X)+\lambda C_H(X)\}],
\label{eq:workflowrisk}
\end{align}
where $\lambda$ expresses the policy trade-off between residual error and review effort. A deployment policy can be posed as
\begin{align}
\min_{t\in\mathcal{T}}\quad & J(t)\\
\text{subject to}\quad & U_g(s,t;\alpha_g)\le\eta_g,\quad g\in\mathcal{G},\\
& \Prob(X\in R_t)\le B,
\label{eq:selectopt}
\end{align}
where $B$ is the maximum review fraction supported by available capacity. If capacity is expressed in reviewer-hours, replace the final constraint with $\E[\1\{X\in R_t\}C_H(X)]\le B_{\mathrm{hours}}$. The feasible threshold set must be determined on a calibration/validation partition, then evaluated on an untouched test partition. If no threshold satisfies all constraints, the system is not qualified for autonomous processing under the stated policy; the threshold must not be relaxed after observing test errors.

The review policy must specify the evidence packet, reviewer competence requirements, permitted actions (confirm, correct, abstain, or escalate), escalation criteria, and audit log. At minimum, evaluation should report review rate, review time, reviewer correction yield, reviewer false-accept rate on sampled machine errors, reviewer false-correction rate on correct outputs, inter-reviewer agreement, and unresolved-case rate. These are proposed estimands; they require a human evaluation study and are not measured in this theoretical paper.

\subsection{A loss-sensitive rule for human review}
Suppose a field has probability $q$ of being correct, an error incurs loss $L>0$, review costs $c\ge0$, and a reviewed field retains residual error probability $r_H$ after review. Expected loss from immediate acceptance is $(1-q)L$, whereas expected loss from review is $c+r_HL$.

\begin{proposition}[Loss-sensitive review condition]
Under this model, review has lower expected loss than immediate acceptance if and only if
\begin{equation}
q < 1-r_H-\frac{c}{L}.
\label{eq:reviewthreshold}
\end{equation}
If $1-r_H-c/L\le0$, review is not preferred on expected-loss grounds for any $q\in[0,1]$ under the assumed costs and residual error rate.
\end{proposition}
\begin{proof}
Review is preferred exactly when $c+r_HL<(1-q)L$. Dividing by $L>0$ and rearranging gives $q<1-r_H-c/L$. If the right-hand side is non-positive, no non-negative $q$ satisfies the strict inequality.
\end{proof}
This result makes explicit that a universal confidence threshold is not generally optimal across fields with different error consequences, review costs, and reviewer effectiveness. The quantities $L$, $c$, and $r_H$ must be estimated or set through a documented policy process. In practice, the model may need to account for heterogeneous reviewer skill, correlated errors, queueing delay, and errors introduced during review.

\begin{proposition}[Monotonicity of coverage]
For any fixed confidence function $c_s$, coverage $\gamma(t)=\Prob(c_s(X)\ge t)$ is non-increasing in $t$.
\end{proposition}
\begin{proof}
For $t_2\ge t_1$, the set $\{X:c_s(X)\ge t_2\}$ is a subset of $\{X:c_s(X)\ge t_1\}$. Taking probabilities yields $\gamma(t_2)\le\gamma(t_1)$.\end{proof}

Selective risk need not be monotone in the threshold without assumptions on the relationship between confidence and loss. Therefore, a risk--coverage curve is an empirical property to estimate, not a theorem implied by thresholding. Calibration must also be assessed at the decision unit: a well-calibrated token score is not automatically a calibrated document-level probability.

\subsection{Annotation uncertainty and reference disagreement}
The reference annotation is itself a measurement. Let $Y^{(1)}$ and $Y^{(2)}$ be independent annotations of the same critical field. Disagreement should be classified as (i) annotator error, (ii) genuinely ambiguous visual evidence, (iii) schema ambiguity, or (iv) source-document inconsistency. An adjudication protocol should retain the original annotations and the adjudicated state. For irreducibly ambiguous fields, the target label should be a set or an explicit state such as illegible/ambiguous, not a fabricated single value. Evaluation should report the proportion of ambiguous fields and the rate at which systems appropriately abstain on them.

\section{Benchmark Specification for Reproducible Evaluation}
\subsection{Annotation schema}
Each document record should contain a stable document identifier, page identifiers, source family, language/script, acquisition metadata, template family, and privacy-preserving provenance. Each region record should contain polygon coordinates, region type, reading-order relation, and annotation status. Each field record should contain field name, document-level entity, verbatim transcription, canonical typed value, unit/currency, sign/polarity, period, table coordinates where relevant, criticality class, evidence-region identifiers, normalization rule identifier, and adjudication status. Missing, illegible, not-applicable, and not-present are distinct states and must not be collapsed into an empty string.

\subsection{Split construction and leakage controls}
The split unit should be the highest practical unit that prevents correlated examples crossing partitions. Depending on the data, this may be customer, account, supplier, issuer, statement, invoice series, template family, or time period. A benchmark should publish at least three evaluation partitions: (i) in-distribution held-out examples from known template families; (ii) template- or institution-held-out examples; and (iii) temporal or acquisition-condition holdouts. The same source document, near-duplicate scan, or template-specific synthetic derivative must not cross partitions. When privacy or licensing prevents release of images, publish hashes, schema, split manifests where permitted, and an executable evaluator with a documented access procedure.

\subsection{Minimum benchmark reporting card}
Table~\ref{tab:benchmarkcard} specifies the minimum information needed to interpret a reported result.
\begin{table*}[!t]
\caption{Minimum reproducibility card for a financial document recognition benchmark}
\label{tab:benchmarkcard}
\centering\footnotesize
\begin{tabularx}{\textwidth}{p{0.20\textwidth}X}
\toprule
Item & Required disclosure \\
\midrule
Population & Target workflow, document families, jurisdictions/languages, sampling frame, collection period, inclusion/exclusion criteria \\
Sampling & Sampling unit, strata, inclusion probabilities, oversampling policy, target deployment weights \\
Ground truth & Annotation manual/version, annotator training, independent double-annotation fraction, adjudication, ambiguity policy, agreement by field type \\
Schema & Field ontology, typed-value representation, units/currencies, polarity, periods, table relations, missingness states, evidence-link rules \\
Partitions & Document/customer/template/time grouping, duplicate detection, holdout rationale, contamination checks \\
Systems & Model and software versions, prompts/configuration, preprocessing, normalization, decoding settings, failure handling \\
Metrics & Metric implementation/version, matching thresholds, micro/macro estimands, criticality weights, gates, confidence procedures \\
Operations & Hardware, concurrency, latency percentiles, throughput, memory, cost boundary, human-review assumptions \\
Governance & Data license/permissions, privacy controls, retention, access controls, benchmark versioning, incident process \\
\bottomrule
\end{tabularx}
\end{table*}

\subsection{Worked decision example (synthetic, not empirical)}
Consider two hypothetical systems evaluated on the same held-out sample. System A has a higher overall normalized score than system B, but its upper confidence bound for polarity errors exceeds the workflow limit. System B has a lower aggregate score, but all critical gates pass. Under Eq.~\ref{eq:eligible}, A is ineligible for automatic processing and B may proceed to the next operational qualification stage. This example demonstrates the logic of the framework only; it does not represent measured performance, an observed dataset, or a recommendation about any commercial system. The evaluator should report, for each system, the point estimate, numerator and denominator, confidence bound, subgroup, and gate decision. A system that fails a gate may still be useful in a human-reviewed workflow, but its auto-accept threshold must be re-evaluated under Eq.~\ref{eq:selectopt}.

\subsection{End-to-end synthetic routing trace}
Consider the fictitious statement in Fig.~\ref{fig:syntheticdocs}. Suppose a document model reads the payroll debit as positive 480,000.00 rather than as a negative outflow. The lexical output may be close to the visible string, but the canonical polarity is wrong. The running-balance validator independently flags the inconsistency. The policy engine records a C3/C4-class gate failure according to the institution's predeclared use case, attaches the source crop and column association, and routes the record to human review. The reviewer may confirm the source transcription, correct the signed value, or mark the source ambiguous; each action is captured with a reason code. The downstream posting adapter remains disabled until the critical gate is resolved.

This trace demonstrates the proposed control logic only. It does not measure detection accuracy, reviewer effectiveness, time saved, financial loss avoided, or production throughput. A real evaluation would compare the framework's routing decisions with adjudicated annotations and downstream outcomes on a governed, representative sample, and would report confidence intervals and review workload alongside any claimed reduction in residual error.
\section{Robustness and Generalization Protocol}
Financial OCR quality must be sliced across acquisition and document conditions. The benchmark should include controlled or naturally occurring variations in resolution, blur, skew, perspective, compression, background noise, faint printing, photocopy generations, stamps, handwriting, overlapping marks, multi-column layout, dense tables, small fonts, mixed scripts, and locale-specific numeric formats. Synthetic perturbations are useful for controlled stress tests but cannot replace naturally degraded samples.

\subsection{Distributional slices}
At minimum, report performance by document category, institution/template, language/script, scan source, page quality, table density, handwriting presence, amount magnitude, negative-value representation, and field criticality. Avoid allowing a large easy category to dominate the aggregate. Let strata be indexed by $h$ with target population proportions $\pi_h$. The target-weighted estimate of a metric is
\begin{equation}
\hat\theta_{\rm strat}=\sum_{h=1}^{H}\pi_h\hat\theta_h,
\qquad \sum_h\pi_h=1.
\end{equation}
If deployment proportions are unknown or unstable, report macro-average across strata and each stratum separately rather than inventing a population mix.

\subsection{Out-of-distribution and template generalization}
Split design should prevent leakage across near-duplicate pages, account holders, vendors, statement periods, and template families where feasible. A random page split can overestimate generalization if pages from the same statement or template occur in both train and test sets. Include temporal holdout and institution/template holdout evaluations. Report performance on known templates and unseen templates separately. If the system is adapted using a test institution's samples, those samples no longer constitute an untouched generalization test.

\section{Statistical Evaluation Design}
\subsection{Sampling and annotation}
The benchmark should be sampled from the intended deployment population using a documented frame. Use stratified sampling to ensure representation of rare but consequential categories. Oversampling rare critical fields is appropriate for diagnostic power, but estimates of population-level performance must be reweighted using known inclusion probabilities. Annotation should be performed by trained annotators using a written schema, with independent double annotation for critical fields and adjudication of disagreements. Report agreement separately for transcription, region localization, canonicalization, and semantic role. The ground-truth process itself is a measurement system and should be audited.

\subsection{Paired system comparison}
When two systems process the same documents, compare document-level paired outcomes. For a scalar document loss $L_d(s)$, define $\Delta_d=L_d(s_A)-L_d(s_B)$. Estimate the mean paired difference $\bar\Delta$ and construct a confidence interval using a document-cluster bootstrap or another method compatible with the sampling design \cite{efron1994}. If multiple pages belong to a document, resample documents, not pages independently. For highly clustered sources, resample at the highest independent sampling unit, such as customer, supplier, or template family. Report effect sizes and confidence intervals, not only p-values.

\subsection{Rare critical errors}
For a critical error probability $p$, an observed zero count in $n$ independent trials does not establish $p=0$. An approximate one-sided 95\% upper bound is $3/n$ (the rule of three) under a binomial model. For exact reporting, use a binomial confidence interval such as Clopper--Pearson \cite{clopper1934}, and account for clustering or unequal sampling where necessary. To demonstrate a target upper bound $p_0$, sample size must be planned before testing; the required size can be large for very small $p_0$. No universal sample size is appropriate without a target error rate, confidence level, design effect, and expected prevalence.

\subsection{Multiplicity and model selection}
If many systems, categories, and metrics are compared, distinguish confirmatory endpoints from exploratory analyses. Pre-register primary metrics and acceptance gates. For multiple confirmatory comparisons, use a prespecified family-wise error procedure such as Holm's step-down method \cite{holm1979}, or control false discovery rate for explicitly exploratory analyses. Do not select the best system on the same test set and then report its unadjusted confidence interval as if the choice had been prespecified. Maintain a final sealed test set for model selection-independent confirmation.

\subsection{Benchmark protocol}
A reproducible evaluation run should record: dataset version and licensing; sampling frame and inclusion probabilities; document hashes; annotation schema version; system/model version; inference parameters; preprocessing; hardware; batch size; concurrency; output schema; normalization rules; metric implementation version; random seeds; and all exclusions. Failed, timed-out, or malformed outputs must be counted and categorized rather than silently dropped. Report both quality conditional on successful output and end-to-end success rate.

\section{Operational Evaluation: Cost, Latency, and Throughput}
Accuracy alone does not determine enterprise suitability. Measure page-level and document-level latency at p50, p90, p95, and p99; throughput under stated concurrency; CPU/GPU and memory consumption; failure rate; retry rate; and cost per 1,000 pages. Separate preprocessing, OCR inference, post-processing, validation, and human-review time. For systems with asynchronous queues, report queueing delay and service time separately. Cost comparisons must use equivalent hardware, batch settings, image resolution, and output requirements, or clearly disclose differences.

Define end-to-end cost per successfully processed document as
\begin{equation}
C_{\rm success}=\frac{C_{\rm inference}+C_{\rm storage}+C_{\rm review}+C_{\rm retry}}{N_{\rm successful}}.
\end{equation}
This metric should be reported with its accounting boundary. A cheaper OCR engine may be more expensive overall if it creates more manual review or downstream correction. Operational reporting should therefore include the auto-accept rate, review rate, correction rate, and cost of residual errors where estimable.

\section{Composite Scorecard and Decision Framework}
A composite score can support concise comparison, but should remain subordinate to the quality vector and hard gates. Let $q_k\in[0,1]$ be normalized dimension scores and $w_k\ge0$ with $\sum_k w_k=1$. A weighted score is
\begin{equation}
S_{\rm weighted}=\sum_{k=1}^{K}w_kq_k.
\end{equation}
This is transparent but compensatory. A geometric alternative,
\begin{equation}
S_{\rm geo}=\prod_{k=1}^{K}\left(\frac{q_k+\varepsilon}{1+\varepsilon}\right)^{w_k},
\end{equation}
penalizes very low dimensions more strongly, but depends on the offset $\varepsilon$ and still does not encode absolute safety requirements. I recommend reporting both only as secondary summaries, alongside all component scores and non-compensatory gates. Weights must be elicited from the intended use case, published before evaluation, and sensitivity-tested over plausible alternatives.

\begin{table*}[!t]
\caption{Proposed enterprise scorecard: dimensions, primary metrics, and mandatory diagnostics}
\setlength{\tabcolsep}{2.5pt}
\label{tab:scorecard}
\centering\footnotesize
\begin{tabularx}{\textwidth}{p{0.15\textwidth}p{0.28\textwidth}X}
\toprule
Dimension & Primary metrics & Required diagnostics \\
\midrule
Text fidelity & CER, WER & Micro/macro; punctuation and symbol policy; critical-token accuracy \\
Detection/\allowbreak localization & Precision, recall, F1 at IoU thresholds & Text-line/word/table region slices; missed-region rate \\
Layout/order & Pairwise reading-order F1; layout-class F1 & Multi-column and cross-page continuation; header/footer confusion \\
Tables & Cell F1; row/column/span accuracy; header association & Cell text exactness; numeric column assignment; merged cells \\
Fields/\allowbreak facts & Typed exact match; precision/recall/F1 & Amount, date, currency, sign, entity, period, identifier slices \\
Risk & Critical error rate; severity-weighted loss & Error counts by C0--C4; upper confidence bounds; gate violations \\
Evidence & Evidence coverage and precision & Unsupported extraction and ungrounded inference rates \\
Uncertainty & Brier score; ECE; risk--coverage & Calibration by category and criticality; review rate \\
Robustness & Degradation under stress; unseen-template performance & Quality strata; language and scan condition slices \\
Operations & p50/p95/p99 latency; throughput; cost/1,000 pages & Success/failure rate; retry; human review; resource profile \\
\bottomrule
\end{tabularx}
\end{table*}

\subsection{Recommended decision tiers}
The framework supports four deployment decision tiers:
\begin{enumerate}[leftmargin=*]
\item \textit{Research/benchmark only}: quality is measured, but one or more critical gates are not met or confidence intervals are too wide.
\item \textit{Human-assisted}: outputs are presented with evidence and uncertainty; a human verifies designated critical fields before downstream use.
\item \textit{Restricted automation}: only prespecified low-risk document categories or fields are auto-accepted; uncertain and critical cases are escalated.
\item \textit{Production automation within scope}: all category-specific gates, statistical assurance requirements, monitoring controls, and operational criteria are satisfied for a defined use case.
\end{enumerate}
These tiers are evaluation outcomes, not certifications. Production scope must name the document population, fields, downstream actions, and operating conditions covered.

\section{Implementation Blueprint for Enterprise Systems}
A practical programme can be implemented in six workstreams.
\begin{enumerate}[leftmargin=*]
\item \textbf{Document inventory and risk mapping.} Catalogue document families, source channels, languages, critical fields, downstream consumers, and error consequences. Map each field to a criticality class and validation rule.
\item \textbf{Ground-truth engineering.} Build a versioned annotation schema with polygons, verbatim text, typed values, semantic relations, evidence links, and ambiguity states. Double-annotate critical fields and adjudicate.
\item \textbf{System adapter and reproducible harness.} Normalize outputs from commercial OCR, open-source engines, layout-aware systems, and OCR-free models into one contract. Freeze versions and inference settings.
\item \textbf{Metric and validator service.} Compute lexical, spatial, structural, field, fact, risk, calibration, and operational metrics. Keep source inconsistency separate from extraction error.
\item \textbf{Statistical reporting.} Produce stratified estimates, confidence intervals, paired comparisons, rare-event bounds, and subgroup reports. Seal the final test set.
\item \textbf{Production monitoring.} Monitor document mix, image quality, template drift, field-level confidence, abstention, correction rates, and critical incidents. Trigger re-evaluation when the input distribution or system version changes.
\end{enumerate}

\begin{figure*}[!t]
\centering
\begin{tikzpicture}[font=\small, node distance=6mm and 7mm,
 stepbox/.style={draw,rounded corners,align=center,minimum height=9mm,text width=27mm,fill=blue!5},
 outputbox/.style={draw,rounded corners,align=center,minimum height=9mm,text width=31mm,fill=green!6},
 arr/.style={-{Latex[length=2mm]},thick}]
\node[stepbox] (ingest) {Ingest and classify\\document + source metadata};
\node[stepbox,right=of ingest] (extract) {Recognize\\text, regions, tables};
\node[stepbox,right=of extract] (normalize) {Normalize\\typed fields + provenance};
\node[stepbox,right=of normalize] (validate) {Validate\\arithmetic + relations};
\node[outputbox,below=of validate] (route) {Confidence and risk routing\\auto-accept / review / reject};
\node[outputbox,left=of route] (audit) {Audit and monitoring\\logs, drift, correction};
\node[outputbox,left=of audit] (consume) {Controlled downstream\\posting / indexing / workflow};
\draw[arr] (ingest)--(extract); \draw[arr] (extract)--(normalize); \draw[arr] (normalize)--(validate);
\draw[arr] (validate)--(route); \draw[arr] (route)--(audit); \draw[arr] (audit)--(consume);
\draw[arr,dashed] (audit.north) to[bend left=18] node[right,align=left,font=\scriptsize]{feedback for\\monitoring only} (extract.south);
\end{tikzpicture}
\caption{Illustrative enterprise processing and control flow. Feedback from monitoring may trigger re-evaluation or controlled system updates; it must not silently rewrite the sealed benchmark labels}
\label{fig:workflow}
\end{figure*}

\subsection{Production control plane, fail-safe states, and interfaces}
The six workstreams above should be deployed as a versioned control plane rather than as an unstructured sequence of scripts. A minimal production record should preserve (i) immutable source-document identity and cryptographic hash; (ii) parser, OCR/model, prompt, preprocessing, and schema versions; (iii) field-level outputs and source-region references; (iv) normalization and validation rule identifiers; (v) risk-policy version and threshold; (vi) routing decision and reason codes; and (vii) reviewer action, timestamp, and subsequent correction. The evaluation service should be logically independent of the recognition service, and benchmark labels should remain access-controlled and unavailable to the production inference path.

The processing interface should be idempotent: retries for the same document and processing version must not create duplicate downstream postings. A document should move through explicit states such as \texttt{RECEIVED}, \texttt{CLASSIFIED}, \texttt{EXTRACTED}, \texttt{VALIDATED}, \texttt{AUTO\_ELIGIBLE}, \texttt{HUMAN\_REVIEW}, \texttt{REJECTED}, and \texttt{COMPLETED}. State transitions should be event logged. An invalid schema, missing evidence for a critical field, failed arithmetic/relationship validator, unsupported document family, timeout, or unrecognized currency/unit should fail closed to review or quarantine; it should not silently fall through to automatic processing. A retry policy should be bounded, and exhausted retries should enter a dead-letter queue with a stable incident identifier.

The downstream adapter should expose only policy-approved fields and should keep the raw visual transcript, canonical value, and business interpretation separately. Posting, payment release, customer communication, or regulatory filing should remain behind the consuming application's existing authorization and approval controls. The OCR framework is an evidence and routing layer; it is not a substitute for maker-checker controls, transaction authorization, segregation of duties, or legal review. Personally identifiable and account data require encryption in transit and at rest, least-privilege access, retention limits, and auditable access to source images. Synthetic examples in this paper contain no real customer or account data.

A staged release should begin in shadow mode, where the system generates outputs without affecting downstream actions. It should then progress to human-assisted use, restricted automation for narrowly defined low-risk fields, and only later to a bounded production scope after independent test-set gates are met. Canary rollout, version pinning, rollback to the last qualified model/policy pair, and a kill switch for critical incidents are required operational controls. Changes to the model, OCR engine, prompt, image preprocessing, schema, normalization, validator, or risk threshold should create a new qualification version; a model-only version number is insufficient because the whole decision chain affects the outcome.

Monitoring should combine leading indicators (document mix, image quality, template novelty, unsupported-field rate, queue depth, timeout rate) with lagging indicators from labelled audits (critical-field error bounds, reviewer correction yield, false-accept rate, unresolved-case rate, and downstream correction/incident rate). Unlabelled drift alerts trigger sampling and review, not a declaration of degraded accuracy. The monitoring plan should define owners, alert thresholds, response times, escalation paths, and rollback authority before production launch. Thresholds should be selected from service and risk requirements, not inferred from the synthetic examples in Fig.~\ref{fig:syntheticdocs}.

\subsection{Monitoring and requalification}
A model update, OCR engine version change, preprocessing change, new document template, or material shift in language or scan quality should trigger a defined requalification policy. Monitoring should distinguish input drift from performance drift. Unlabelled drift indicators can trigger sampling and review but cannot establish accuracy. Periodic labelled audits are needed to estimate current error rates. Critical errors should have incident handling, root-cause classification, and corrective-action tracking.

\section{Threats to Validity and Limitations}
First, ground truth is not infallible. Ambiguous handwriting, poor scans, source inconsistencies, and differing accounting conventions can create annotation disagreement. The framework therefore requires explicit ambiguity states and adjudication rather than forcing a single answer. Second, risk weights are context-dependent. A universal weighting scheme would create false precision; institutions must map errors to their own workflows and controls. Third, consistency rules may be invalid when the document uses special accounting treatments, rounding, reversals, or jurisdiction-specific conventions. Validators must be scoped and versioned.

Fourth, benchmark data may not represent future deployment distributions. Template holdouts and temporal tests reduce, but do not eliminate, this risk. Fifth, aggregate scores can conceal subgroup failures; all critical slices should be reported. Sixth, confidence calibration can deteriorate after model or distribution changes. Seventh, an OCR benchmark cannot by itself establish that a downstream financial action is legally or operationally appropriate. Human authorization, access control, audit trails, and transaction controls remain outside the OCR metric layer.

Finally, the framework is proposed theoretically and methodologically. It has not been empirically validated in this paper; the worked decision example is synthetic and is included only to clarify the acceptance logic. The proposed human-review estimands, evidence-packet requirements, and loss-sensitive routing rule are design specifications, not findings from a user study or deployment. The paper does not release a benchmark dataset, annotation manual validated by annotators, software implementation, or measured reviewer-performance data. A future study should instantiate the benchmark across multiple financial document families, evaluate representative OCR and document AI systems, measure annotation cost and inter-annotator agreement, and test whether risk-aware metrics better predict downstream correction and incident rates than CER/WER alone.

\section{Research Agenda}
Several empirical questions follow directly. (i) Does criticality-weighted evaluation correlate more strongly with downstream correction cost than lexical error? (ii) Which evidence-grounding protocol is reliable across OCR-dependent and OCR-free models? (iii) How well do token- and field-level confidence estimates predict critical financial errors? (iv) What sample sizes are needed to bound rare critical errors under realistic clustering? (v) How should multi-document consistency be evaluated for invoices, purchase orders, remittance advice, and bank statements? (vi) Can a common benchmark schema support multilingual and jurisdiction-specific financial formats without erasing local conventions? These questions define a testable programme rather than an unsupported claim of a solved problem.

\section{Conclusion}
Financial OCR must be evaluated according to the information and relationships it preserves, not only the characters it reproduces. This paper proposed a layered evaluation framework spanning visual recognition, transcription, layout, tables, typed fields, financial facts, evidence grounding, uncertainty, robustness, and operational performance. It formalized criticality-weighted loss while emphasizing non-compensatory gates for high-impact fields; defined risk--coverage analysis for selective automation; and specified a statistically defensible benchmark protocol that accounts for stratification, paired comparisons, clustering, rare events, and multiplicity. The framework is intended to be implementable in enterprise document-processing systems and adaptable across financial document categories. Its central principle is that high average OCR accuracy is not sufficient evidence of financial reliability: critical values must be correct, structurally associated, visually grounded, and appropriately routed when uncertain.

\section*{Statements and Declarations}
\subsection*{Funding}
No external funding was received for this research.

\subsection*{Competing Interests}
The author declares no competing financial or non-financial interests related to this work.

\subsection*{Author Contributions}
Subodh Lonkar: Conceptualization, methodology, formal analysis, investigation, writing---original draft, and writing---review and editing. The author reviewed and approved the submitted version and accepts responsibility for the integrity of the work.

\subsection*{Data Availability}
No empirical dataset was collected, generated, or analysed in this theoretical study. The four fictitious document exemplars are included in Fig.~\ref{fig:syntheticdocs}; their individual image files and deterministic rendering script are supplied in the source archive. Public datasets discussed in the manuscript are identified by their original publications and should be accessed from their respective repositories subject to applicable licences and terms of use.

\subsection*{Code Availability}
No implementation of the proposed evaluation framework was developed or evaluated. The deterministic script used to render the four illustrative synthetic documents is included in the source archive.

\subsection*{Ethics Approval}
Not applicable. This theoretical study involved no human participants, animals, identifiable personal data, or intervention.

\subsection*{Consent to Participate and Consent to Publish}
Not applicable.

\subsection*{Broader Impact}
The proposed framework may improve the transparency and control of automated financial-document processing by making critical-field failures, evidence provenance, and human-review decisions measurable. Misuse is also possible: poorly chosen risk thresholds, unrepresentative benchmarks, or automation pressure could create false assurance, shift workload unfairly to reviewers, or allow incorrect financial records to propagate. Mitigations include workflow-specific criticality policies, independent test sets, uncertainty bounds, fail-closed routing, human override, audit logging, privacy controls, and periodic labelled requalification. The framework does not certify regulatory compliance or authorize financial transactions.

\subsection*{Use of Generative AI and AI-Assisted Tools}
ChatGPT (OpenAI) was used during manuscript preparation to assist with research planning, conceptual organization, literature discovery, drafting and revision of explanatory text, and language refinement. The author reviewed and takes responsibility for the final manuscript, including the verification of mathematical statements, citations, interpretations, arithmetic in examples, and claims. AI-generated suggestions were not treated as empirical evidence. The four illustrative document panels in Fig.~\ref{fig:syntheticdocs} were created programmatically from manually specified fictitious text and numerical values using a deterministic Python/ReportLab script; they are not photographs, scans of real records, or generative-AI images.

\end{document}